\documentclass[final]{cvpr}

\usepackage{times}
\usepackage{epsfig}
\usepackage{graphicx}
\usepackage{amsmath}
\usepackage{amssymb}
\usepackage{multirow}
\usepackage{bigstrut}
\usepackage{booktabs}
\usepackage{comment}
\usepackage{bbm}

\usepackage[pagebackref=true,breaklinks=true,colorlinks,bookmarks=false]{hyperref}

\def\cvprPaperID{6406} 
\def\confYear{CVPR 2021}

\begin{document}

\title{Co-Grounding Networks with Semantic Attention for Referring Expression Comprehension in Videos}

\author{Sijie Song{$\small^{1}$}, Xudong Lin{$\small^{2}$}, Jiaying Liu{$\small^{1}$}, Zongming Guo{$\small^{1}$}\thanks{Corresponding author.},~~Shih-Fu Chang{$\small^{2}$}\\
$\small ^{1}$\ Wangxuan Institute of Computer Technology, Peking University, Beijing, China\\
$\small ^{2}$\ DVMM Lab, Columbia University, New York, NY, USA
}

\maketitle

\begin{abstract}
In this paper, we address the problem of referring expression comprehension in videos, which is challenging due to complex expression and scene dynamics. Unlike previous methods which solve the problem in multiple stages (i.e., tracking, proposal-based matching), we tackle the problem from a novel perspective, \textbf{co-grounding}, with an elegant one-stage framework. We enhance the single-frame grounding accuracy by semantic attention learning and improve the cross-frame grounding consistency with co-grounding feature learning. Semantic attention learning explicitly parses referring cues in different attributes to reduce the ambiguity in the complex expression. Co-grounding feature learning boosts visual feature representations by integrating temporal correlation to reduce the ambiguity caused by scene dynamics. Experiment results demonstrate the superiority of our framework on the video grounding datasets VID and LiOTB in generating accurate and stable results across frames. Our model is also applicable to referring expression comprehension in images, illustrated by the improved performance on the RefCOCO dataset. Our project is available at \url{https://sijiesong.github.io/co-grounding}.


\end{abstract}

\section{Introduction}
Referring expression comprehension has attracted much attention recently. It aims to localize a region of the image/video described by the natural language. This topic is of great importance in computer vision to support a variety of research problems such as image/video captioning~\cite{anderson2018bottom,pan2017video}, visual question answering~\cite{antol2015vqa} and image/video retrieval~\cite{vo2019composing,gabeur2020multi}. It also plays a key role in machine intelligence for a wide range of applications from human-computer interaction, robotics to early education.

\begin{figure}[t] 
	\begin{center}
		\includegraphics[width=1.0\linewidth]{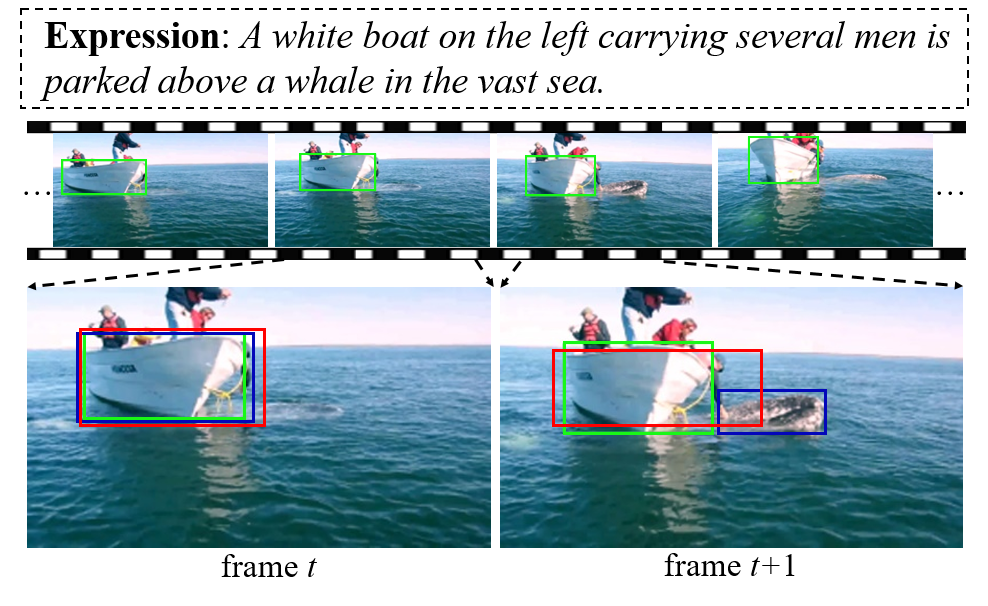}
	\end{center}
\vspace{-4mm}
	\caption{Referring expression comprehension in videos. Due to dynamic scenes and ambiguity in the expression, per-frame inference (in \textcolor{blue}{blue}) with state-of-the-art grounding method~\cite{yang2019fast} would lead to unstable results across frames, while our co-grounding networks achieve accurate and consistent predictions (in \textcolor{red}{red}). Ground-truth annotations are denoted in \textcolor{green}{green}.}
	\label{fig:teaser}
\vspace{-6mm}
\end{figure}

In the past years, most of the previous work for referring expression comprehension focus on the grounding for static images~\cite{wang2019neighbourhood,yang2019cross,yu2018mattnet,liao2020real,yang2019fast,mao2016generation,nagaraja2016modeling,hu2017modeling,akbari2019multi} and have achieved promising results. However, referring expression comprehension for videos is less explored, which is challenging yet important. Different from several threads of referring expression comprehension in videos, such as referring all mentioned entities~\cite{zhou2018weakly}, we localize the spatio-temporal tube that semantically corresponds to the whole sentence. That is, we output bounding box for each frame as shown in Figure~\ref{fig:teaser}.

The work in~\cite{li2017tracking} treats referring expression comprehension for videos as a tracking problem. We argue that it would suffer from template selection error, because it is hard to tell from the grounding results from multiple frames which is the right one to track. The other work~\cite{chen2019weakly} first proposes spatio-temporal tube candidates and then matches them with textual features from expression. However, the performance is limited by the proposal quality. Inspired by the advance in one-stage image grounding methods~\cite{yang2019fast} that get rid of proposal detectors, another solution is to conduct per-frame inference with~\cite{yang2019fast}, but there are still two problems. Firstly, the entities in the expression (such as `\emph{boat}', `\emph{men}', `\emph{whale}', `\emph{sea}' in Figure~\ref{fig:teaser}) would cause ambiguity when encoded into textual features, making the model confused about which entity is the correct one to ground. Secondly, the dynamic scenes across frames would also interrupt the grounding process (see the drifting blue bounding boxes in Figure~\ref{fig:teaser}). Therefore, the key challenge is to generate robust textual and visual features to reduce ambiguity, then further achieve accurate and stable results across frames.



To tackle the aforementioned issues, we propose to solve referring expression comprehension in videos with a new perspective, \emph{i.e.}, co-grounding, with semantic attention learning in an elegant one-stage framework. The basic structure of our model is based on YOLO~\cite{redmon2018yolov3}, which predicts the bounding box and confidence simultaneously. The confidence reflects the matching score between the textual and visual features. We design a semantic attention mechanism to obtain attribute-specific features both for vision and language. Specifically, a proposal-free subject attention scheme is proposed to parse the words for subject from the expression. An object-aware location attention scheme is developed to parse the words for location from the expression. The interaction between attribute-specific textual and visual features determines the subject score and location score for each visual region (see the visualization examples in Figure~\ref{fig:vis_attn}). Besides, to improve cross-frame prediction consistency, we develop the co-grounding feature learning. Taking multiple frames as input, it utilizes the correlation across frames to enhance visual features and stabilize the grounding process in training and testing. A post-processing strategy is further employed to improve the temporal consistency during inference.

Our contributions are summarized as follows:

$\bullet$ We propose to solve referring expression comprehension in videos by co-grounding in an one-stage framework.

$\bullet$ We propose semantic attention learning to parse referring cues, including a proposal-free subject attention and object-aware location attention.

$\bullet$ Our networks are applicable to both video/image grounding, and achieve state-of-the-art performance on referring expression comprehension benchmarks.



\section{Related Work}
\subsection{Referring expression comprehension}

Benefiting from the advances in object detection~\cite{ren2015faster,redmon2018yolov3,he2017mask}, most methods~\cite{wang2019neighbourhood,yang2019cross,yu2018mattnet} perform referring expression comprehension in two stages. Region candidates are proposed in the first stage with an object detector and then matched with the expression in the second stage. The best matching region is selected as the grounding result. However, these two-stage methods are limited by the proposal quality of the offline object detector. The missing of ground-truth regions in the first stage would lead to the failure of the second stage. To address the issue, more recent works are proposed to get rid of the offline object detector~\cite{liao2020real,yang2019fast}. Built upon the current one-stage object detection method, \emph{i.e.}, YOLO-v3~\cite{redmon2018yolov3}, Yang \emph{et al.}~\cite{yang2019fast} first proposed an one-stage visual grounding framework, which extracts visual-text features and predict bounding boxes densely at all spatial locations. Liao \emph{et al.}~\cite{liao2020real} reformulate referring expression comprehension as correlation filtering, where the filter template is generated from language features. Besides referring expression comprehension for images, there are a few works~\cite{chen2019weakly,li2017tracking} exploring the task in the video domain. However, both of the methods solve referring expression comprehension for videos with multiple stages (\emph{i.e.}, tracking, proposal-based matching). The failure in the first stage would directly impact the final results. In our work, we propose an elegant one-stage framework for this task.

\begin{figure*}[t] 
	\begin{center}
		\includegraphics[width=1.0\linewidth]{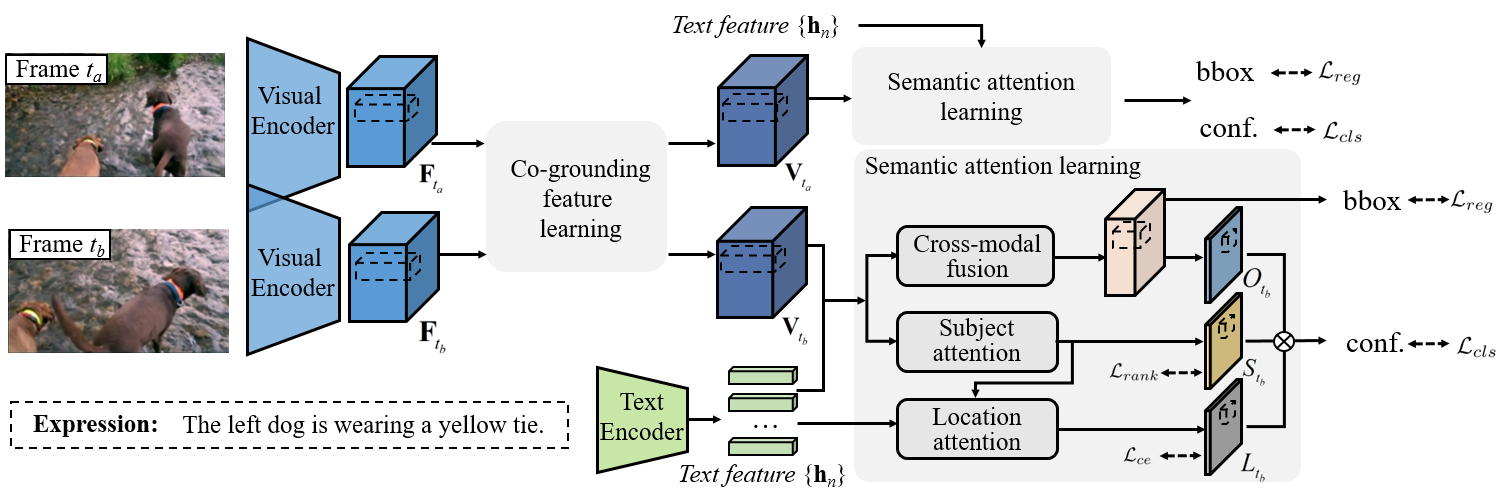}
	\end{center}
\vspace{-3mm}
	\caption{Our co-grounding networks with semantic attention for referring expression comprehension. The details for semantic attention learning and co-grounding feature learnng are presented in Figure~\ref{fig:semantic_attention} and Figure~\ref{fig:cogrounding}, respectively.}
\vspace{-3mm}
	\label{fig:framework}
\end{figure*}

\subsection{Attention mechanisms}
Inspired by human perception, attention mechanisms have been widely studied in vision-and-language tasks, \emph{e.g.}, visual question answering~\cite{yu2019deep,lu2016hierarchical,kim2020hypergraph,nguyen2018improved}, visual dialogue~\cite{wu2018you} and visual grounding~\cite{yu2018mattnet}. In these works, attention is applied to learn the underlying correlation between different modalities~\cite{yu2019deep,lu2016hierarchical}, which jointly performs language-guided visual attention and vision-guided language attention. Nguyen \emph{et al.}~\cite{nguyen2018improved} propose a dense symmetric co-attention to deal with every interaction between any pair of visual region and each word. To further fully understand the semantic of vision and language, a more recent work~\cite{kim2020hypergraph} presents a hypergraph model to define a common semantic space among different modalities. Attention mechanisms are also popular in the task of visual grounding to decompose the language into several components~\cite{yu2018mattnet}. Each component focuses different attributes of the language and then trigger corresponding visual comprehension. However, the modular network designed in~\cite{yu2018mattnet} is based on offline object proposals and requires external annotations. Our work, however, achieves semantic attention learning without the reliance of object proposals and external labels.


\subsection{Temporal consistency}
Generating consistent results across adjacent frames is essential for video applications.The recent works mainly focus on improving performance of per-frame result by exploiting information in the temporal domain~\cite{feichtenhofer2017detect,zhu2017flow,chen2020memory,li2018recurrent,lu2019see,hu2018relation,nilsson2018semantic,chen2017coherent}. Some of the works aggregate local temporal context to help the inference of current frame. Optical flow is always computed by~\cite{dosovitskiy2015flownet} to propagate features across frames~\cite{zhu2017flow,chen2017coherent}, while some methods integrate temporal context by calculating affinity matrix~\cite{feichtenhofer2017detect,lu2019see} to build temporal correspondence. Nevertheless, only focusing the locality may lead to the lack of long-term information. To aggregate information beyond a small local range,~\cite{chen2020memory} introduces a global-local aggregation network, taking full consideration of both global and local information. In addition,~\cite{li2018recurrent,nilsson2018semantic} share a similar idea which leverages the merits of recurrent networks to make use of neighboring results. In our work, we design a co-grounding module to leverage correlation across frames, and further stabilize the prediction results with a post-processing strategy.


\section{The Proposed Method}
Given an expression $Q$ with $N$ words and a video $\mathbf{I}$ with $T$ frames, our goal is to localize the object region $\{b_t\}_{t=1}^T$ in each frame described by $Q$, where $b_t$ represents a bounding box in the $t$-th frame.

We build our baseline model following~\cite{yang2019fast}, which is based on the one-stage object detection framework, \emph{i.e.}, YOLOv3~\cite{redmon2018yolov3}. We conduct bounding box prediction based on cross-modal features, which are obtained by fusing visual and textual features. For each spatial position of the cross-modal features, the model outputs bounding box predictions centered at the current spatial position, with a confidence to indicate the probability of being the final grounding output. The bounding box with the highest confidence is selected as the final prediction. The basic objective function for training the model consists of the MSE (mean square error) loss $\mathcal{L}_{reg}$ to regress the bounding box towards the ground-truth, and a cross-entropy loss $\mathcal{L}_{cls}$ to select the right prediction from all the bounding boxes. We refer readers to~\cite{yang2019fast, redmon2018yolov3} for more details. Next, we elaborate the semantic attention learning and co-grounding feature learning introduced to the framework.


\subsection{Semantic attention learning}
To reduce ambiguity from the expression, we propose semantic attention learning to parse referring cues from the input expression. Though the input expression is usually complex, it is noticed that the words indicating subject and location play a key role to distinguish the target. Thus, we aim to decompose the expression into subject and location. With more attribute-specific textual features, we build the mapping between language and vision. Note that our semantic attention learning is different from~\cite{yu2018mattnet} since our networks are end-to-end trainable without the reliance of offline proposal detection and external label annotations.


As shown in Figure~\ref{fig:framework}, the expression $Q$ is encoded with a text encoder consisting of bi-directional LSTM. The representation for the $n$-th word is the concatenation of the hidden states from both directions:
\vspace{-1mm}
\begin{equation}
\label{equ:exp_encoder}
\mathbf{h}_n = [\overrightarrow{\mathbf{h}}_n,\overleftarrow{\mathbf{h}}_n] = \text{BiLSTM}(\mathbf{e}_n, \overrightarrow{\mathbf{h}}_{n-1},\overleftarrow{\mathbf{h}}_{n+1}),
\end{equation}
where $\mathbf{e}_n$ is the embedding of the $n$-th word. The attribute-specific textual features $\mathbf{q}_m$ ($m \in \{sub., loc.\}$) are parsed by fusing $\{\mathbf{h}_n\}_{n=1}^N$ with learnable weights $\mathbf{w}^m \in \mathbb{R}^N$:


\begin{equation}
\label{equ:semantic_attn}
{\alpha}_n^m = \frac{\text{exp}\left(w^m_n \mathbf{h}_n\right)}{\sum_{i=1}^N\text{exp}\left(w^m_i \mathbf{h}_i\right)},
\end{equation}
\begin{equation}
\label{equ:weighted_exp}
\mathbf{q}_m = \sum_{n=1}^N {\alpha}_n^m \mathbf{e}_n.
\end{equation}

\begin{figure}[t] 
	\begin{center}
		\includegraphics[width=1.0\linewidth]{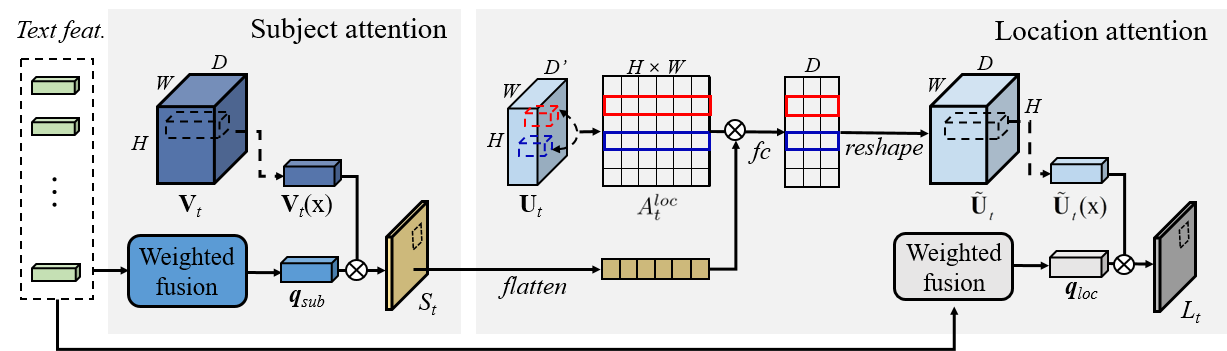}
	\end{center}
\vspace{-2mm}
	\caption{Details for semantic attention learning.}
	\label{fig:semantic_attention}
\vspace{-4mm}
\end{figure}

$\bullet$ \textbf{Proposal-free subject attention.} In this part, our networks learn $\mathbf{w}^{sub}$ to parse subject from $Q$ and generate subject attention map $S_t$ for each frame in a proposal-free manner. The subject attention map $S_t \in \mathbb{R}^{H \times W}$ reflects visual feature response to $\mathbf{q}_{sub}$ by computing cross-modal similarity as shown in Figure~\ref{fig:semantic_attention}:
\begin{equation}
\label{equ:score_subject}
S_t(x) = \delta(\mathbf{V}_t(x), \mathbf{q}_{sub}) = \mathbf{q}_{sub}[{\mathbf{V}_t(x)}]^{\mathbf{T}},
\end{equation}
where $\mathbf{V}_t(x) \in \mathbb{R}^D$ is the feature vector at the position $x$ from visual feature map $\mathbf{V}_t \in \mathbb{R}^{H \times W \times D}$ of the $t$-th frame. Rank loss is exploited to train the network, where the score of the matched visual and textual features, \emph{i.e.}, ($\mathbf{V}_t(x^*), \mathbf{q}_{sub}$), should be higher than unmatched ones, \emph{i.e.}, ($\mathbf{V}_t(x^*), \mathbf{q}'_{sub}$) and ($\mathbf{V}_t(x'), \mathbf{q}_{sub}$). Therefore, the objective function is:
\begin{equation}
\begin{aligned}
\label{equ:rank_loss}
\mathcal{L}_{rank} & =  \max\left(0, \Delta + \delta\left(\mathbf{V}_t(x'),\mathbf{q}_{sub}\right) - \delta\left(\mathbf{V}_t(x^*),\mathbf{q}_{sub}\right)\right)  \\
& +  \max\left(0, \Delta + \delta\left(\mathbf{V}_t(x^*),\mathbf{q}'_{sub}\right) - \delta\left(\mathbf{V}_t(x^*),\mathbf{q}_{sub}\right) \right),
\end{aligned}
\end{equation}
where $\Delta$ is a margin and set to $0.5$ in our experiments. During training, the visual feature vector corresponding to $\mathbf{q}_{sub}$ can be localized with the ground-truth annotation. The key issue in the training is how to select negative visual feature vectors without having object proposals. Though different schemes of hard negative sample mining have been explored, we found it is enough to tackle the problem by random sampling visual features from other training samples within the same training batch.

$\bullet$ \textbf{Object-aware location attention.} In this part, our networks learn $\mathbf{w}_{loc}$ to parse location from the expression, and then generate location map $L_t \in \mathbb{R}^{H \times W}$ to $\mathbf{q}_{loc}$. The key challenge in learning location attention is how to match coordinates with textual features because location is a relative concept. When we say something is `\emph{on the left}', we have to give a reference. Thanks to the aforementioned subject attention map $S$ which roughly identifies the subject region, we design an object-aware location representation. Specifically, we follow~\cite{yang2019fast} to initially encode the coordinate feature as $\mathbf{U}_t \in \mathbb{R}^{H \times W \times D'}$. A 2D matrix $A^{loc}_t \in \mathbb{R}^{{HW}\times{HW}}$ is computed to model the relation between any two positions $x$ and $y$:
\begin{equation}
\label{equ:loc_affinity}
A^{loc}_t(x,y) = [\mathbf{U}_t(x)]^{\mathbf{T}} \mathbf{U}_t(y).
\end{equation}
With the subject attention map $S_t$, we inject the reference information into the location features as $A^{loc}_t \otimes \text{Flatten}(S_t)$, followed by an FC layer to shape it into $HW \times D$. Then the matrix is reshaped to $H\times W \times D$ as the final location features $\tilde{\mathbf{U}}_t$. A detailed illustration is shown in Figure~\ref{fig:semantic_attention}.

Similar to Eq.~\ref{equ:score_subject}, we obtain the location response for position $x$ by computing cosine similarity $L_t(x) = \delta(\tilde{\mathbf{U}}_t(x), \mathbf{q}_{loc})$.
We train the location attention with cross-entropy loss:
\begin{equation}
\label{equ:ce_loss}
\mathcal{L}_{ce} = \mathbbm{1}_{loc} \log \frac{\exp(L_t(x))}{\sum_{y}\exp(L_t(y))},
\end{equation}
where $\mathbbm{1}_{loc} \in \{0,1\}^{H\times W}$ indicates the ground-truth location.

With the subject and location attention maps, the confidence map for the $t$-th frame is generated as $C_t = O_t \otimes S_t \otimes L_t$. Recall that we generate a bounding box prediction for each position $x$, $O_t(x)$ indicates how likely the predicted bounding box contains an object.

\begin{figure}[t] 
	\begin{center}
		\includegraphics[width=1.0\linewidth]{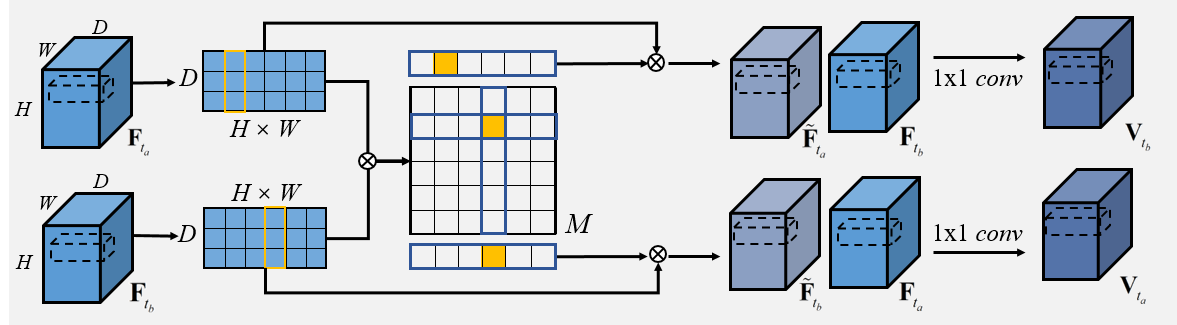}
	\end{center}
\vspace{-2mm}
	\caption{Details for co-grounding feature learning.}
\vspace{-4mm}
	\label{fig:cogrounding}
\end{figure}

\subsection{Co-grounding feature learning}
For now, we have introduced how to parse referring cues from expression and build correspondence between text and visual features. However, the temporal dynamics in videos would lead to unstable visual feature representations, which may do harm to the cross-modal matching in both training and testing. To enhance the visual feature representation for a more robust learning, we propose co-grounding to integrate the temporal context by utilizing correlation across frames. As shown in Figure~\ref{fig:framework}, considering two frames from the same video, we obtain the initial visual features $\mathbf{F}_{t_a} \in \mathbb{R}^{H \times W \times D}$ and $\mathbf{F}_{t_b} \in \mathbb{R}^{H \times W \times D}$ with a visual encoder. The correlation across the adjacent frames can be described by a normalized affinity matrix $M \in \mathbb{R}^{HW \times HW}$, providing the measure for similarity of spatial features:
\begin{equation}
\label{equ:norm_affinity_mat}
M(x,y)=  \frac{\exp\left( [\mathbf{F}_{t_a}(x)]^\mathbf{T} \mathbf{F}_{t_b}(y) \right)}{\sum_y \exp\left( [\mathbf{F}_{t_a}(x)] ^\mathbf{T}  \mathbf{F}_{t_b}(y) \right)}.
\end{equation}

Then we integrate the feature vectors from $\mathbf{F}_{t_b}$ with M,
\begin{equation}
\label{equ:f_tb}
\tilde{\mathbf{F}}_{t_b}(x) = \sum_{y} M(x,y)\mathbf{F}_{t_b}(y).
\end{equation}
The final enhance feature $\mathbf{V}_{t_a}$ is obtained by:
\begin{equation}
\label{equ:enhanced_feat_va}
\mathbf{V}_{t_a} = \text{Conv}(\mathbf{F}_{t_a} \oplus \tilde{\mathbf{F}}_{t_b}),
\end{equation}
where $\oplus$ indicate concatenation along the channel dimension, and $\text{Conv}(\cdot)$ denotes an $1\times 1$ convolution operation. The visual feature $\mathbf{V}_{t_b}$ can be enhanced in the same way. Figure~\ref{fig:cogrounding} shows the details of co-grounding feature learning.

\subsection{Post processing}
To further stabilize the bounding box prediction for each frame, we design a post-processing scheme based on the initial prediction results during inference. Suppose for the video, we have the initial top $K$ bounding box predictions for each frame $\{ \{b^i_t, c^i_t\}_{i=1}^K \}_{t=1}^T$, where $b^i_t$ denotes the location for the bounding box with the $i$-th highest confidence $c^i_t$ for the $t$-th frame. The visual feature vectors corresponding to the bounding box location for the $t$-th frame is $\mathcal{V}_{t} = \mathbf{V}^1_{t}(b^1_{t})\oplus ...\oplus \mathbf{V}^K_{t}(b^K_{t})$ and $\mathcal{V}_{t} \in \mathbb{R}^{D\times K}$. Now we consider referring the neighboring $P$ frames as a window to stabilize the center frame ${t^*}$. For the $i$-th bounding box of the center frame, we stabilize its confidence score by seeking the most similar bounding box in each reference frame. The similarity is measured by the affinity matrix:
\begin{equation}
\label{equ:post_processing_affinity}
Z_{(t^*,t)} = \mathcal{V}_{t^*}^{\mathbf{T}} \mathcal{V}_t, t = \{t^* - \Delta t, ..., t^* + \Delta t\},
\end{equation}
where $\Delta t = \lfloor P/2 \rfloor$, and $Z_{(t^*,t)} \in \mathbb{R}^{K \times K}$. For each bounding box in $\{b_{t^*}^i\}_{i=1}^K$, we select the most similar bounding box from all the reference frames,
\begin{equation}
\label{equ:select_idx}
\mathbf{p}_t = [p^{(1)}_t, ..., p^{(K)}_t],
\end{equation}
where
\begin{equation}
\label{equ:detail_idx}
p^{(i)}_t = \arg \max Z^{(i)}_{(t^*,t)},
\end{equation}
and $Z^{(i)}_{(t^*,t)}$ denotes the $i$-th row of the matrix.
The final scores for the initial top $K$ bounding boxes are:
\begin{equation}
\label{equ:final_score}
\tilde{\mathcal{C}}_{t^*} = \frac{1}{\mathcal{N}}\sum_{t=t^*-\Delta t}^{t^* + \Delta t} \mathbbm{1}_{\mathbf{p}_t} * \mathcal{C}_t,
\end{equation}
where $\tilde{\mathcal{C}}_{t^*} \in \mathbb{R}^{K}$, $\mathcal{C}_t \in \mathbb{R}^{K}$ and $\mathcal{C}_t = [c_t^1, ..., c_t^K]$. $\mathbbm{1}_{\mathbf{p}_t} \in [0,1]^K$ can be regard as a binary mask to choose the highest score from $\mathcal{C}_t$. The bounding box with the highest score in $\mathcal{C}_{t^*}$ is treated as the final prediction result. $\mathcal{N}$ is a normalization factor.

\begin{table*}[htbp]
  \begin{center}
    \caption{Referring expression comprehension results on dynamic video datasets VID and LiOTB, respectively.}
    \label{table:video_results}
    \begin{tabular}{l c c c c c c}
    \toprule
              &\multicolumn{3}{c}{VID}  & \multicolumn{3}{c}{LiOTB} \\
              \cmidrule(r){2-4} \cmidrule(r){5-7}
              & Accu.@0.5 & Success & Precision & Accu.@0.5 & Success & Precision \\
             \hline
             One-Stage BERT~\cite{yang2019fast}        & 52.39 & 0.427 & 0.373 & 49.13  &   0.358    &   0.468   \\
             One-Stage LSTM~\cite{yang2019fast}        & 54.78 & 0.451 & 0.393 & 49.16 &   0.333    &   0.414   \\
             First-Frame Tracking  & 36.97 & 0.334 & 0.250 & 50.93 & 0.391 & 0.482\\
             Middle-Frame Tracking & 44.00 & 0.384 & 0.307 & 43.08 & 0.356 & 0.421\\
             Last-Frame Tracking   & 36.26 & 0.328 & 0.239 & 44.17 & 0.327 & 0.391\\
             Random-Frame Tracking & 40.20 & 0.356 & 0.278 & 25.16 & 0.288 & 0.329\\
             WSSTG~\cite{chen2019weakly}               & 38.20 & - & - & -  &  -   &  -    \\
             LSAN~\cite{li2017tracking}                & - & - & - & - & 0.259 & - \\
             \hline
             Ours & \textbf{60.25} & \textbf{0.495} & \textbf{0.462} & \textbf{52.26} & \textbf{0.392} & \textbf{0.500} \\
     \bottomrule
    \end{tabular}
  \end{center}
  \vspace{-5mm}
\end{table*}

\section{Experiments}
In this section, we introduce the datasets and implementation details, then report our evaluation on referring expression comprehension benchmarks. We first show the comparisons to other state-of-the-art methods, then give the ablation study and comprehensive analysis to illustrate the effectiveness of each component. Finally we discuss the failure cases and future work.

\subsection{Datasets}
To evaluate our model, we conduct experiments on two dynamic video datasets (\emph{i.e.}, VID-Sentence~\cite{chen2019weakly}, Lingual OTB99~\cite{li2017tracking}) and one static image dataset (\emph{i.e.}, RefCOCO).

\textbf{VID-Sentence (VID)}~\cite{chen2019weakly}. This dataset consists of 7,654 trimmed videos with language descriptions, and provides the sequences of spatio-temporal bounding box annotations for each query. Following~\cite{chen2019weakly}, the dataset is splited into 6,582/536/536 instances for training/validation/testing.

\textbf{Lingual OTB99 (LiOTB)}~\cite{li2017tracking}. The LiOTB dataset origins from the well-known OTB100 object tracking dataset in~\cite{lu2014online}. The videos in~\cite{lu2014online} are augmented with natural language descriptions of the target object. We adopt the same protocol as~\cite{li2017tracking} that 51 videos are for training and the rest are for testing.

\textbf{RefCOCO}~\cite{yu2016modeling}. The RefCOCO dataset is collected from 19,994 images in MSCOCO~\cite{lin2014microsoft} and 142,210 natural language descriptions. RefCOCO is splited into four subsets, including train, validation, test A and test B. The images in test A are with multiple people, while those in test B are with multiple objects.

\subsection{Implementation details}
\noindent \textbf{Training settings.} Our visual encoder is based on Darknet-53~\cite{redmon2018yolov3} pretrained on MSCOCO~\cite{lin2014microsoft}. We adopt multi-level schemes in the grounding process, that we predict bounding boxes on three levels of feature maps, the resolution of which are $8 \times 8$, $16 \times 16$ and $32 \times 32$. The input images are resized the long edges to 256 and then padded into the size of $256 \times 256$. Following~\cite{redmon2018yolov3,yang2019fast}, we adopt the data augmentation including adding randomization to the color space, horizontal flip, and random affine transformations. The network is optimized with RMSProp~\cite{tieleman2012lecture} for 100 epochs, the initial learning rate of which is set as $10^{-4}$ and decayed under a polynomial schedule. The batch sizes for VID, LiOTB and RefCOCO are 32, 8, 32, respectively. We set the weights for $\mathcal{L}_{rank}$ and $\mathcal{L}_{ce}$ as 100, 1, respectively. By default, the top 5 bounding boxes from the neighboring 5 frames are considered in the post-processing for the VID and LiOTB datasets during inference (\emph{i.e.}, $K=5$, $P=5$). Note for the image dataset RefCOCO, we omit the co-grounding feature learning.

\noindent \textbf{Evaluation metrics.} We adopt different metrics to give a fair and comprehensive evaluation of our framework. Acc@0.5 is widely used to evaluate the grounding results~\cite{yang2019fast,liao2020real}, where a predicted bounding box is considered correct if the IoU with the ground truth region is above 0.5. Following~\cite{yang2019grounding}, success and precision scores are reported to evaluate the performance for videos. The success score is actually the AUC (area under curve) metric, while the precision score measures the ratio of frames where the predicted bounding box falls within a threshold of 20 pixels around the ground-truth. Besides, mIoU is also reported to show the quality of bounding boxes.

\subsection{Comparison to the state-of-the-art}

Table~\ref{table:video_results} shows the referring expression comprehension results on the video datasets VID and LiOTB, respectively. We first present the per-frame inference results by the state-of-the-art referring expression comprehension method~\cite{yang2019fast} in the first two rows. From the 3rd to the 6th rows, we evaluate the results to see the performance when the problem is solved by per-frame tracking. Specifically, we adopt the state-of-the art tracker~\cite{li2019siamrpn++} to track the given template. With the grounding results from One-Stage LSTM~\cite{yang2019fast}, we conduct experiments with the first, middle, last and random frame as the tracking template, respectively. It is found that treating referring expression grounding in videos as tracking leads to poor results compared to per-frame grounding for almost all the cases. There are mainly two reasons to explain the poor results. On the one hand, we can not guarantee the correct template is selected during tracking. On the other hand, even given the correct template, the tracker may fail due to its limitation. The analysis also applies to the unsatisfactory results of LSAN~\cite{li2017tracking}. Besides, we present the results of WSSTG~\cite{chen2019weakly} which relies on spatio-temporal proposal detection. Our results are shown in the last row. It can be seen our framework outperforms the compared methods by a large margin.


\begin{table}[t]
    \small
  \begin{center}
    \caption{Referring expression comprehension results (Acc.@0.5) on the static image dataset RefCOCO.}
    \label{table:refcoco_results}
    \begin{tabular}{l c c c c}
    \toprule
              &\multicolumn{4}{c}{RefCOCO} \\
              \cmidrule(r){2-5}
              & Visual encoder & val & testA & testB \\
             \hline
             MMI~\cite{mao2016generation} & VGG16 & - & 64.90 & 54.51 \\
             Neg Bag~\cite{nagaraja2016modeling} & VGG16 & - & 58.60 & 56.40 \\
             CMN~\cite{hu2017modeling} & VGG16 & - & 71.03 & 65.77  \\
             SLR~\cite{yu2017joint} & ResNet-101 & 69.48 & 73.71 & 64.96 \\
             DGA~\cite{yang2019dynamic} & ResNet-101 & - & 78.42 & 65.53 \\
             MAttN~\cite{yu2018mattnet} & ResNet-101 & 76.40 & 80.43 & 69.28 \\
             CMCF~\cite{liao2020real} & DLA-34 & - & 81.06 & 71.85 \\
             One-Stage BERT~\cite{yang2019fast} & Darknet53 & 72.05 & 74.81 & 67.59\\
             One-Stage LSTM~\cite{yang2019fast} & Darknet53 &73.69 & 75.78 & 71.32 \\
             \hline
             Baseline & Darknet53 & 73.72 & 76.24 & 71.19 \\
             S-Att. & Darknet53 & 77.42 & \textbf{81.17} & 72.77\\
             SL-Att.(Ours) & Darknet53 & \textbf{77.65} & 80.75 & \textbf{73.37} \\
     \bottomrule
    \end{tabular}
  \end{center}
  \vspace{-8mm}
\end{table}

Table~\ref{table:refcoco_results} shows the comparisons on the RefCOCO dataset for referring expression comprehension. Our overall results on RefCOCO are shown in the last row.
To give a fair comparison, we present the backbone structure of each visual encoder. Though MAttN~\cite{yu2018mattnet} also explores the idea of modular attention for both language and vision, it requires an offline object detector and external attribute labels. However, our model can learn the semantic attention in a proposal-free manner, which also makes our results outstanding compared to other one-stage models~\cite{liao2020real,yang2019fast}.

\begin{table}[t]
  \begin{center}
    \caption{Referring expression comprehension results for ablation study on dynamic video datasets VID and LiOTB, respectively.}
    \label{table:ablation}
    \small
    \begin{tabular}{l c c c c}
    \toprule
              &\multicolumn{2}{c}{VID}  & \multicolumn{2}{c}{LiOTB} \\
              \cmidrule(r){2-3} \cmidrule(r){4-5}
             & Acc.@0.5 & mIoU & Acc.@0.5 & mIoU \\
             \hline
             \multicolumn{5}{l}{\emph{w/o co-grounding}} \\
             \hline
             Parser & 53.19 & 0.450 & 49.16 & 0.397\\
             Baseline & 54.41 & 0.448 & 49.11 & 0.405 \\
             S-Att. & 58.03 & 0.488 & 49.66 & 0.411\\
             SL-Att. & \textbf{59.22} & \textbf{0.490} & \textbf{50.51} & \textbf{0.418}\\
             \hline
             \multicolumn{5}{l}{\emph{w/ co-grounding}} \\
             \hline
             CG-Baseline & 55.88 & 0.477 & 50.56 & 0.405 \\
             CG-S-Att.  & 58.74 & 0.497&  51.67 & 0.412\\
             CG-SL-Att. & 59.48& 0.494 &50.92 & 0.418 \\
             CG-SL-Att. + pp. & \textbf{60.25} & \textbf{0.498} & \textbf{52.26}&\textbf{0.418}\\
     \bottomrule
    \end{tabular}
  \end{center}
    \vspace{-8mm}
\end{table}

\subsection{Ablation study}
\vspace{-2mm}
To show the effectiveness of each component in our model, ablation study is conducted on VID, LiOTB and RefCOCO datasets, respectively. We explore different settings to give a comprehensive analysis. The results are presented in Table~\ref{table:ablation} and Table~\ref{table:refcoco_results}. Note that we regard the model of one-stage LSTM in~\cite{yang2019fast} as our \textbf{Baseline}.

 \begin{figure*}[t] 
	\begin{center}
		\includegraphics[width=1.0\linewidth]{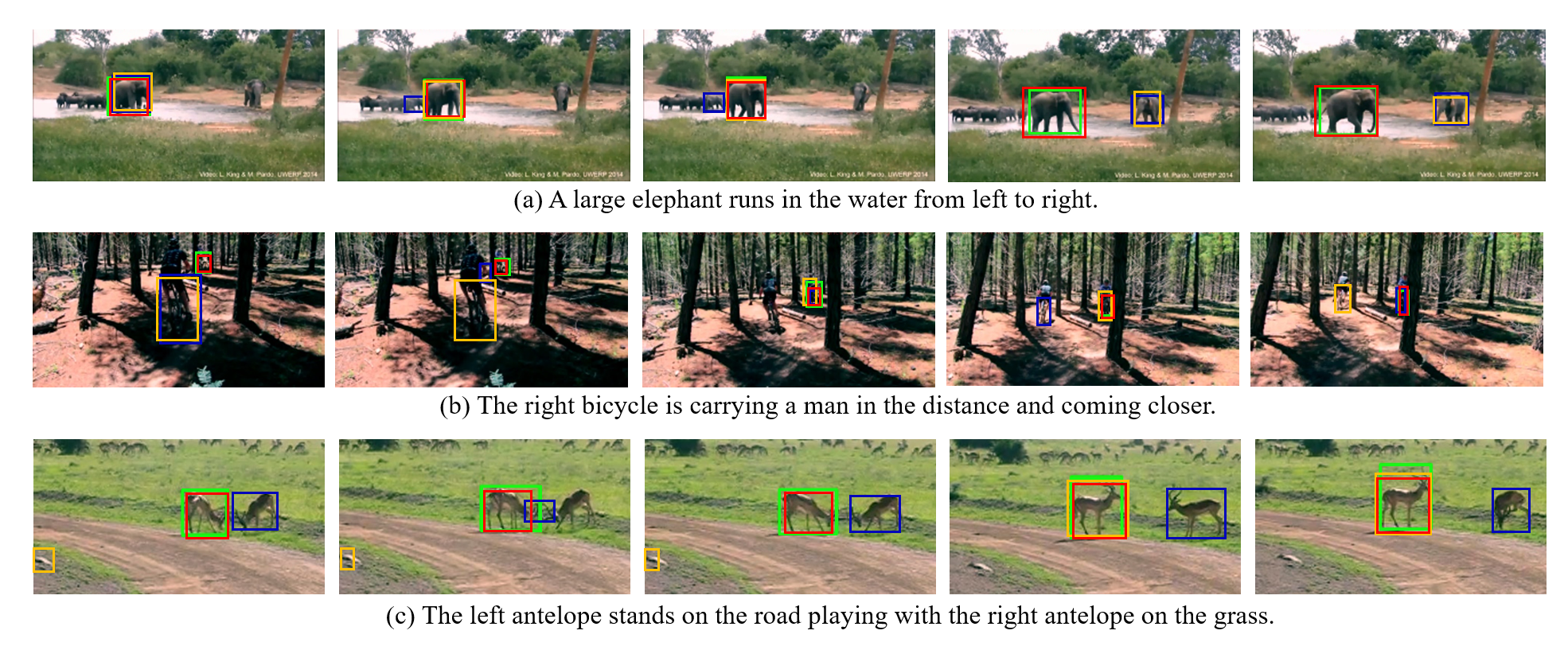}
	\end{center}
\vspace{-5mm}
	\caption{Visualization results of video grounding on the VID dataset. We show the \textcolor{green}{ground-truths}, \textcolor{blue}{baseline results}, \textcolor[rgb]{1,0.7,0}{results of SL-Att.} and \textcolor{red}{results of CG-SL-Att.} in the green, blue, orange and red bounding boxes, respectively. The language queries are shown in the sub-captions.}
	\label{fig:vis_grounding}
\vspace{-4mm}
\end{figure*}

$\bullet$ \textbf{Semantic attention learning.} We first explore the contribution of semantic attention learning without taking co-grounding feature learning into account.
For all the datasets, subject attention (\textbf{S-Att.}) brings significant improvement compared to baseline results. It is largely because the subject attention reduces ambiguity in grounding process when there are multiple entities in the expression and images. Location attention (\textbf{SL-Att.}) further improves the grounding accuracy. Overall, for the VID and LiOTB datasets, the gains from semantic attention learning over baselines are 4.81\% and 1.40\%, respectively. For the RefCOCO dataset, our framework outperforms the baseline by 3.93\%, 4.93\%, and 2.18\% under different split settings, respectively. Moreover, we compare our automatic semantic attention learning with manually semantic parser~\cite{kazemzadeh2014referitgame} in Table~\ref{table:ablation} (see $\textbf{Parser}$). As analyzed in~\cite{yu2018mattnet}, parsing errors exist for the external parser which is not tuned for referring expressions. Therefore, we did not observe improvement compared to baseline results.

$\bullet$ \textbf{Co-grounding feature learning.} We further analyze the contributions of our co-grounding feature learning through Table~\ref{table:ablation}. Conducting co-grounding feature learning with the baseline structure (\textbf{CG-Baseline}) brings 1.47\% and 1.45\% gains in terms of Acc.@0.5 over \textbf{Baseline} on the VID and LiOTB datasets, respectively. With semantic attention learning, the co-grounding feature learning helps further improve the performance on both datasets for each setting (see \textbf{CG-S-Att.} \emph{vs.} \textbf{S-Att.}, \textbf{CG-SL-Att.} \emph{vs.}  \textbf{SL-Att.}).

\begin{table}[t]
  \begin{center}
    \caption{Ablation study for post-processing on the VID dataset in terms of Acc.@0.5 and mIoU.}
    \label{table:post_processing}
    \begin{tabular}{l c c c c c}
    \toprule
                            & {P=1} & {P=3} & {P=5} & {P=7} & {P=9} \\
    \hline
    Acc.@0.5                &  59.48 &  59.50 & 60.25 & 60.16 & 59.63  \\
    \hline
    mIoU                    &  0.494 & 0.495 & 0.500 & 0.498 & 0.495 \\
     \bottomrule
    \end{tabular}
    \vspace{-9mm}
  \end{center}
\end{table}

$\bullet$ \textbf{Post processing.} Finally we illustrate the effectiveness of the post processing scheme. The results in the last row of Table~\ref{table:ablation} (\textbf{CG-SL-Att. + pp.}) show that our post-processing scheme is able to further improve the grounding results in terms of Acc.@0.5 and mIoU. In Table~\ref{table:post_processing}, we further explore the influence of different numbers of reference frames. While the post-processing scheme consistently improve the results compared to those without post-processing (\emph{i.e.}, $P=1$), we set $P$ as 5 in all our experiments for the tradeoff between efficiency and accuracy.

\begin{figure*}[htbp] 
	\begin{center}
		\includegraphics[width=1.0\linewidth]{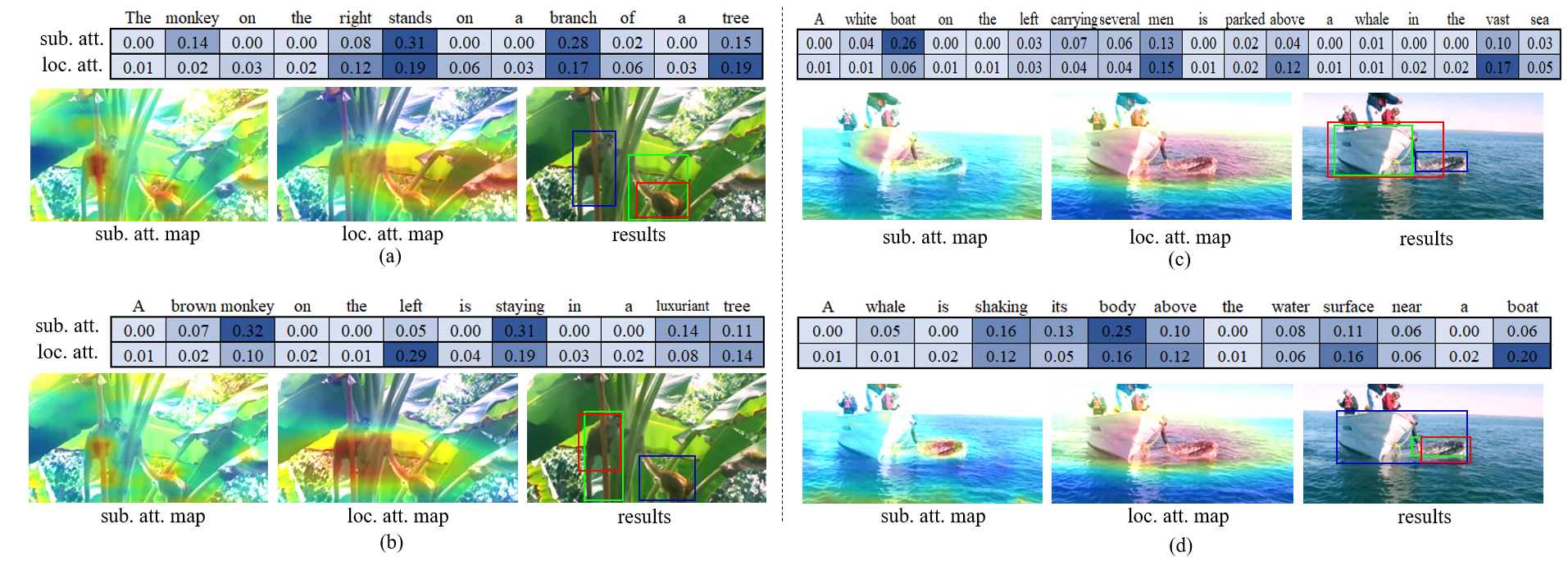}
	\end{center}
\vspace{-6mm}
	\caption{Visualization on attention patterns of our framework. The subject and location attention patterns for language are shown above the images. We mark the attention values for each word. The attention patterns for visual features are shown as the heat maps overlayed on the original images. The more reddish, the larger attention. Besides, we show the grounding results of \textcolor{blue}{baseline} and \textcolor{red}{SL-Att.} in the blue and red bounding boxes, respectively. \textcolor{green}{Ground-truths} are shown in green. (Best viewed in color.)}
	\label{fig:vis_attn}
\vspace{-4mm}
\end{figure*}

\subsection{Qualitative results}
\vspace{-2mm}
$\bullet$ \textbf{Overall grounding results.} We choose several videos from the VID dataset and visualize the grounding results in Figure~\ref{fig:vis_grounding}. We compare the results of the baseline, SL-Att. and CG-SL-Att. with the blue, orange and red bounding boxes, respectively. The ground-truths are in green. From the visualization, it is observed that compared to baseline results, SL-Att. provides more accurate prediction in most cases, due to the explicitly parsed referring cues both for language and vision. However, bounding box drifting is a problem when we conduct per-frame inference without taking the temporal context into account (see the drifting orange bounding boxes). In Figure~\ref{fig:vis_grounding}(b), the target bicycle is small and obscure in the first several frames, making the visual features vulnerable for SL-Att. model. In Figure~\ref{fig:vis_grounding}(c), the left stone mislead SL-Att. to ground it as `\emph{antelope}'. With co-grounding feature learning, the visual features are enhanced by integrating temporal context and become more robust. Therefore, we obtain consistent results across frames (see the red boxes). Please refer to the supplementary for more grounding results.


\vspace{-2mm}
$\bullet$ \textbf{Visualization on attention.} We show the learned attention patterns for language and vision in Figure~\ref{fig:vis_attn}. For each side, different queries are given for the same frame. It is found that the semantic attention for the given expression successfully parses the words for subject entities and location from the language. And the semantic attention for the visual feature maps generate corresponding response for different attributes. In Figure~\ref{fig:vis_attn}(a) and (b), we show examples that location attention helps the model handle ambiguities in the frames and distinguish the correct bounding box. With the parsed subject `\emph{monkey stands branch}' and `\emph{monkey staying}', the model pays more attention on both monkeys in the frame. However, with the guidance of the parsed words describing location `\emph{right stands branch tree}' and `\emph{left staying}', the model shows different response on the location attention maps, providing essential cues to distinguish the correct bounding box. For the examples on the right side, there is not dominant location information in the input expressions, resulting in similar location maps in Figure~\ref{fig:vis_attn}(c) and (d). The multiple entities such as `\emph{whale}', `\emph{boat}' appearing in the sentences make the baseline model confusing that which subject is correct to ground (see the blue bounding boxes in Figure~\ref{fig:vis_attn}(c)(d)). In our model, the subject attention for language effectively excludes the effect of other entities in the expression, making it clear to ground `\emph{boat}' in Figure~\ref{fig:vis_attn}(c) and `\emph{body}' in Figure~\ref{fig:vis_attn}(d). It further leads to corresponding high response on the subject attention maps for visual features and then satisfactory grounding predictions (see the red boxes). More visualizations can be found in our supplementary.

\begin{figure}[t] 
	\begin{center}
		\includegraphics[width=1.0\linewidth]{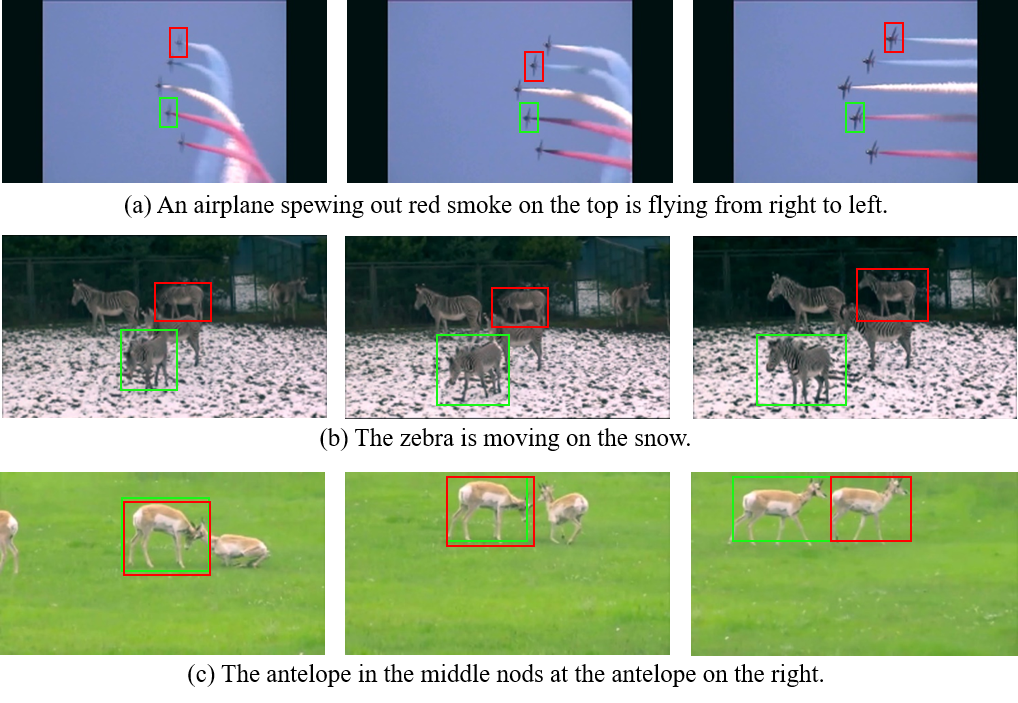}
	\end{center}
\vspace{-5mm}
	\caption{Failure cases. \textcolor{green}{Ground-truths} are in green and \textcolor{red}{our results} are in red.}
\vspace{-7mm}
	\label{fig:failure_case}
\end{figure}

$\bullet$ \textbf{Failure case analysis and future work.}
We show some typical failure cases in Figure~\ref{fig:failure_case}, to illustrate the limitations our model, and the challenges for the topic of video grounding. (1) Multi-order reasoning is challenging for one-stage referring expression comprehension because it always involves multiple entities and relation concepts. As shown in Figure.~\ref{fig:failure_case}(a), it is difficult for our model to locate the airplanes with red smoke and then select the top one. (2) Motion information is not explicitly explored and utilized as grounding cues. As shown in Figure~\ref{fig:failure_case}(b), we can not determine which `\emph{zebra}' is `\emph{moving}' only by observing static frames. (3) The language query may not apply to all the frames. In Figure~\ref{fig:failure_case}(c), the ground-truth is not in the `\emph{middle}' in some frames. How to further tackle the ambiguity caused by dynamic scenes and expression is still worth exploring. We leave how to solve these failure cases as interesting future works.

\section{Conclusion}
In this paper, we tackle the problem of referring expression comprehension in videos. We propose to solve the problem from a new perspective, co-grounding, with an elegant one-stage framework. To boost single frame results, our model learns semantic attention to decompose grounding cues into different attributes, which further contribute to the reasoning of the target described by the input expression. To boost grounding prediction consistency, we propose co-grounding feature learning by integrating neighboring features across frames to enhance visual feature representations. A post-processing scheme is conducted during inference to further stabilize the predictions. Our model is applicable to visual grounding both for videos and images. Experiments on video and image grounding benchmarks illustrate the effectiveness of our model.

\small \noindent\textbf{Acknowledgements.} This work was supported by the Beijing Natural Science Foundation under Contract No.4192025, the National Natural Science Foundation of China under Contract No.61772043. This is a research achievement of Key Laboratory of Science, Technology and Standard in Press Industry (Key Laboratory of Intelligent Press Media Technology). This work was done in part when Sijie Song was a visiting scholar at Columbia University, supported by China Scholarship Council.

{\small
\bibliographystyle{ieee_fullname}
\bibliography{cvpr2021_arxiv}
}

\end{document}